**Tipografía Time New Roman 12 ptos.        Interlineado a 1.5      Justificación completa para el cuerpo**

# Sistema experto para el diagnóstico de enfermedades y plagas en los cultivos del arroz, tabaco, tomate, pimiento, maíz, pepino y frijol.

## Expert system for the diagnosis of diseases and pests in rice, tobacco, tomato, pepper, corn, cucumber and bean crops.

Ing. Yosvany Medina Carbó [1*] https://orcid.org/0000-0003-3590-0706

MSc. Iracely Milagros Santana Ges[1] https://orcid.org/0000-0003-4588-1318

Lic. Saily Leo González[1] https://orcid.org/0000-0002-0804-891X

[1] Centro Universitario Municipal ¨Hermanos Saíz Montes de Oca¨ Consolación del Sur. Calle 53 No 6808 e/ 68 y 70 Consolación del Sur, Pinar del Río, Cuba.

*Autor para la correspondencia: yosvany.medina@upr.edu.cu

**RESUMEN**

La producción agrícola se ha convertido en un negocio complejo que requiere la acumulación y la integración de conocimiento, además de la información de muchas fuentes diversas. Para permanecer competitivo, el agricultor moderno a menudo confía en especialistas agrícolas y consejeros que les proporcionan información para la toma de decisiones en sus cosechas. Pero lamentablemente, la ayuda del

**Tipografía Time New Roman 12 ptos.          Interlineado a 1.5     Justificación completa para el cuerpo**

especialista agrícola no está siempre disponible cuando el agricultor la necesita. Para aliviar este problema, los sistemas expertos se han convertido en un instrumento poderoso y que posee un gran potencial dentro de la agricultura. Este trabajo presenta un Sistema Experto para el diagnóstico de enfermedades y plagas en los cultivos del arroz, tabaco, tomate, pimiento, maíz, pepino y frijol. Para el desarrollo de este Sistema Experto se utilizó SWI-Prolog para la creación de la base de conocimientos por lo que trabaja con predicados y permite que el sistema se base en reglas de producción. Este sistema permite realizar un diagnóstico rápido y fiable de las plagas y enfermedades que afectan a estos cultivos.

**Palabras clave**: cultivos; diagnóstico; plagas y enfermedades; sistema experto; toma de decisiones.

**ABSTRACT**

Agricultural production has become a complex business that requires the accumulation and integration of knowledge, in addition to information from many different sources. To remain competitive, the modern farmer often relies on agricultural specialists and advisors who provide them with information for decision making in their crops. But unfortunately, the help of the agricultural specialist is not always available when the farmer needs it. To alleviate this problem, expert systems have become a powerful instrument that has great potential within agriculture. This paper presents an Expert System for the diagnosis of diseases and pests in rice, tobacco, tomato, pepper, corn, cucumber and bean crops. For the development of this Expert System, SWI-Prolog was used to create the knowledge base, so it works with predicates and allows the system to be based on production rules. This system allows a fast and reliable diagnosis of pests and diseases that affect these crops.

**Tipografía Time New Roman 12 ptos.**     **Interlineado a 1.5**     **Justificación completa para el cuerpo**

# Introducción

La Agenda 2030 para el Desarrollo Sostenible, que incluye 17 Objetivos y 169 metas, presenta una visión ambiciosa del desarrollo sostenible e integra sus dimensiones económica, social y ambiental. Esta nueva Agenda es la expresión de los deseos, aspiraciones y prioridades de la comunidad internacional para los próximos 15 años. En su objetivo 2: Poner fin al hambre, lograr la seguridad alimentaria y la mejora de la nutrición y promover la agricultura sostenible se plantea que si se hace bien, la agricultura, la silvicultura y las piscifactorías pueden suministrarnos comida nutritiva para todos y generar ingresos decentes, mientras se apoya el desarrollo de las gentes del campo y la protección del medio ambiente. Dentro de sus metas para el 2030 este objetivo manifiesta poner fin al hambre y asegurar el acceso de todas las personas, en particular los pobres y las personas en situaciones de vulnerabilidad, incluidos los niños menores de 1 año, a una alimentación sana, nutritiva y suficiente durante todo el año; poner fin a todas las formas de malnutrición, incluso logrando, a más tardar en 2025, las metas convenidas internacionalmente sobre el retraso del crecimiento y la emaciación de los niños menores de 5 años, y abordar las necesidades de nutrición de las adolescentes, las mujeres embarazadas y lactantes y las personas de edad; duplicar la productividad agrícola y los ingresos de los productores de alimentos en pequeña escala, en particular las mujeres, los pueblos indígenas, los agricultores familiares, los ganaderos y los pescadores, entre otras cosas mediante un acceso seguro y equitativo a las tierras, a otros recursos e insumos de producción y a los conocimientos, los servicios financieros, los mercados y las oportunidades para añadir valor y obtener empleos no agrícolas; asegurar la sostenibilidad de los sistemas de producción de alimentos y aplicar prácticas agrícolas resilientes que aumenten la productividad y la producción, que contribuyan al mantenimiento de los ecosistemas, fortalezcan la capacidad de adaptación al cambio climático, los fenómenos meteorológicos extremos, las sequías, las inundaciones y otros desastres, y mejoren progresivamente la calidad de la tierra y el suelo.(Naciones Unidas, 2016)

**Tipografía Time New Roman 12 ptos.         Interlineado a 1.5      Justificación completa para el cuerpo**

Para dar cumplimiento a estas metas planteadas en el objetivo 2 de la Agenda 2030 para el Desarrollo Sostenible el sector alimentario y el sector agrícola ofrecen soluciones claves para el desarrollo y son vitales para la eliminación del hambre y la pobreza. La seguridad alimenticia global es el balance entre la creciente demanda de alimentos de la población mundial y la producción mundial agrícola, combinada con discrepancias entre suministro y demanda a escalas regionales, nacionales y locales (Savary, y otros, 2012). El estado de la seguridad alimenticia global en la actualidad es alarmante y ha empeorado durante los últimos años, la crisis alimenticia del 2008 ha tenido un efecto devastador en este sentido (FAO, 2015).

La mayor causa de pérdida en la producción agrícola lo constituyen las enfermedades producidas por microorganismos fitopatógenos, tales como bacterias, nematodos u hongos, que provocan grandes pérdidas en los cultivos tanto en cosecha como en post cosecha. Otro de los factores que influyen en la pérdida de alimentos son las plagas, quienes provocan enormes prejuicios tanto a la economía como a la seguridad alimenticia global. Nuestro país no está exento de estas afectaciones, la actividad desarrollada por hongos, bacterias y virus en los órganos de las plantas también origina disminuciones en la calidad y en los rendimientos de los cultivos.

El impacto negativo de estas enfermedades en los cultivos incrementa los costos de producción agrícola, por la necesidad de implementar estrategias adicionales para el control del microorganismo patógeno y la disminución de sus efectos sobre las cosechas. Es por ello que la producción agrícola se ha convertido en un proceso dinámico complejo que requiere la acumulación e integración de conocimiento e información de diversas fuentes a través de modelos o sistemas de apoyo para alertar, monitorear y controlar las diferentes plagas y enfermedades que atacan a los cultivos.

El sector productivo y la tecnología siempre han ido de la mano, dando como resultado increíbles investigaciones, y han aportado grandes avances al desarrollo de sistemas enfocados a apoyar específicas actividades del sector productivo. La Inteligencia Artificial y una de sus ramas, los sistemas expertos, con un enfoque totalmente diferente al resto de los sistemas existentes, brindan la posibilidad de tomar

**Tipografía Time New Roman 12 ptos.        Interlineado a 1.5      Justificación completa para el cuerpo**

decisiones de manera más precisa y rápida de lo que un ser humano lo haría, lo que los hace una herramienta potente dado a que está en la capacidad de procesar grandes volúmenes de información, sin correr el riesgo de tomar una decisión errónea, por no tener en cuenta datos que considere innecesarios.

Los sistemas expertos surgen de las técnicas de la inteligencia artificial que han sido objeto de amplias investigaciones desde el año 1950, pero la investigación en este tema comenzó realmente en los 60's donde surgieron los primeros artículos en este campo. Sánchez y Beltrán, en su trabajo sobre sistemas expertos, establecen una metodología de programación y describen la siguiente definición: "Un sistema experto es un conjunto de programas que son capaces, mediante la aplicación de conocimientos, de resolver problemas en un área determinada del conocimiento o saber que ordinariamente requirieran de la inteligencia humana" (Sánchez, y otros, 2009), mientras que Kandel, define un sistema experto como "un sistema informático que simula los procesos de aprendizaje, memorización, razonamiento, comunicación y acción de un experto humano, en una determinada ciencia, suministrando de esta forma, un consultor que puede sustituirle con unas ciertas garantías de éxito" (Kandel, 1992). Sin embargo, todo esto sugiere que el conocimiento sólo se adquiere con un largo aprendizaje y en base a experiencia, de ahí que un sistema experto deba tener ciertos elementos para su correcta funcionalidad. Se requiere que dicho sistema al ser instalado en un ordenador, sea capaz de resolver, por medio de inferencias, problemas específicos de razonamiento con buenas probabilidades de solución. El desarrollo de los Sistemas Inteligentes (SI) y sus continuos avances han permitido su expansión y aplicación en diversas áreas del conocimiento y tecnología. En el campo de la agricultura se han desarrollado trabajos enfocados al diagnóstico de enfermedades en plantas.

En fitopatología, ciencia que se encarga del estudio de las enfermedades de las plantas, se utilizan con frecuencia estos sistemas con fines de diagnóstico, por ejemplo, identificar la causa de una enfermedad por los síntomas y las observaciones relacionadas. Mediante la incorporación de modelos de infección de enfermedades de cultivos en la base de conocimientos de la computadora, el SE puede asesorar a los productores sobre la probabilidad de ocurrencia real de la enfermedad, períodos de infección y recomendaciones sobre tipo, cantidad y momento de aplicación de pesticidas. (Moreno, 2005) (Bula, y otros, 2013)

**Tipografía Time New Roman 12 ptos.          Interlineado a 1.5     Justificación completa para el cuerpo**

Por la importancia del diagnóstico de las enfermedades en los cultivos y analizando las ventajas de tener un sistema que asista el diagnóstico de plagas y enfermedades que los afectan, nos proponemos desarrollar un sistema experto para realizar el diagnóstico de enfermedades en los cultivos de arroz, tabaco, tomate, pimiento, maíz, pepino y frijol. De esta forma serán tomadas decisiones preventivas y correctivas que ayuden a mejorar la productividad agrícola en el municipio Consolación del Sur.

# Métodos o Metodología Computacional (A neg 16 ptos. Cent.)

En este trabajo se presenta un Sistema Experto para el diagnóstico y detección de plagas y enfermerdades en los cultivos de interés agrícola del municipio Consolación del Sur. Este sistema permite realizar un diagnóstico rápido y fiable de las paglas y enfermedades que afectan a los cultivos de: arroz, tabaco, tomate , maíz, pimiento, pepino y frijol. En esta sección se describe brevemente la metodología utilizada y se detalla su utilización para el desarrollo del sistema propuesto.

## Metodología para la construcción del sistema experto.

El estudio del presente trabajo estuvo conformado por cinco fases independientes pero profundamente relacionadas, que fueron desarrolladas por los autores con ayuda de expertos humanos.

La primera fase fue una investigación previa que consistió en recolectar información concerniente a la construcción de un sistema experto y en hacer un estudio detallado de las herramientas o lenguajes de programación utilizados para la construcción de estos, con la intención de elegir la herramienta más apropiada para dar solución a la problemática.

La segunda fase fue una planeación, la cual tenía como objetivo producir un documento estructurado y organizado que permitiera el desarrollo del Sistema Experto. Para ello se establecieron procesos tales como: determinación de las Tareas del Sistema Experto, definición de los requisitos de alto nivel.

**Tipografía Time New Roman 12 ptos.        Interlineado a 1.5       Justificación completa para el cuerpo**

La tercera fase fue definir la base de conocimiento sobre la cual se tomarían las decisiones que permitieran diagnosticar las plagas y enfermedades en los cultivos de arroz, tabaco, maíz, pimiento, pepino, frijol y tomate mediante su sintomatología, para esto lo primero que se hizo fue adquirir el conocimiento necesario relacionado con el tema objeto de estudio, el cual se obtuvo de revisión de literatura como libros, artículos y revistas, pero principalmente de entrevistas personales con especialistas en estos cultivos.

La cuarta fase fue la de codificación y verificación, esta fase consistió en diseñar las reglas de inferencia que se tendrían en cuenta en la base del conocimiento, concerniente al diagnóstico de las plagas y enfermedades en los cultivos de arroz, tabaco, maíz, pimiento, pepino, frijol y tomate mediante su sintomatología. De otro lado en esta fase también se diseñó la arquitectura del sistema experto para el diagnóstico de las plagas y enfermedades en los cultivos de arroz, tabaco, maíz, pimiento, pepino, frijol y tomate mediante su sintomatología, para integrar todos los módulos del sistema y permitir a los usuarios la interacción con el sistema mediante una interfaz para la toma de decisiones.

La quinta y última fase consistió en evaluar el sistema, esto se logró mediante la realización de una prueba detallada del sistema, sin errores de compilación, para así poder construir el manual del usuario.

# Resultados y discusión

**Sistema experto para el diagnóstico de enfermedades y plagas en los cultivos del arroz, tabaco, tomate, pimiento, maíz, pepino y frijol.**

Uno de los mayores problemas en el sector agrícola es el azote de plagas y enfermedades, llegando a ser muy perjudiciales y generan grandes pérdidas, tanto de cultivos como monetarias. Si bien es cierto, esto genera muchas dificultades para los agricultores, ya que repercute en su trabajo, también se debe tener en cuenta que no todos poseen el conocimiento necesario ni la capacidad para poder afrontar este tipo de

**Tipografía Time New Roman 12 ptos.        Interlineado a 1.5        Justificación completa para el cuerpo**

desastres. En consecuencia, se planteó el desarrollo de un sistema inteligente experto que, gracias a los conocimientos de un agente humano experto, será capaz de diagnosticar, prevenir y tratar a este tipo de problemas para que cualquier persona, y más aún, los agricultores se vean beneficiados al tomar algunas acciones inmediatas que los ayude a combatir estas plagas y enfermedades.

Se buscó a un experto humano que proporcionó la base del conocimiento para el sistema inteligente. Las ideas, experiencia e información de este experto humano sirvieron para que el sistema pueda responder a muchas de las preguntas que un agricultor se plantea al momento que una plaga o una enfermedad azota a su cultivo.

Después de la realización de un análisis intensivo de la toda la información proporcionada por el experto humano y por la bibliografía consultada, se determinó usar las reglas de producción en lenguaje Prolog, por adaptarse mejor a las condiciones del problema. Para ello la organización del conocimiento quedó dividida en 7 módulos, uno por cada cultivo. Cada uno de estos módulos dispone de un predicado principal que será el responsable de invocar al conocimiento almacenado en esa parte.

El sistema permite realizar el diagnóstico de diferentes tipos de plagas y enfermedades en 7 cultivos. En la Tabla 1 se aprecia cada uno de los cultivos y los tipos de enfermedades y plagas que los puede afectar.

**Tabla 1** Plagas y enfermedades diagnosticados por el sistema experto

| Cultivo | Plagas y enfermedades |
|---|---|
| Arroz | ✓ Piricularia (Pyricularia oryzae), Chilo Supressalis o Barrenador del Arroz, Pyricularia grisea del arroz, Rosquillas, Pudenta (Eysarcoris ventralis) |
| Tabaco | ✓ Pythium aphanidermatum (Damping off), Peronopora hyoscyami (Moho azul del tabaco), Phytophthora (Pata prieta) |
| Tomate | ✓ Araña Roja (Tetranychus urticae), Mosca Blanca (Trialeurodes vaporariorum y Bemisia tabaci) , Minador (Liriomyza trifolii, Liriomyza bryoniae, Liriomyza strigata y Liriomyza huidobrensis), Polilla (Tuta absoluta), Mildiu (Phytophthora infestans), Podredumbre gris (Botrytis cinerea), Cladosporiosis (Fulvia fulva), |

**Tipografía Time New Roman 12 ptos.     Interlineado a 1.5     Justificación completa para el cuerpo**

|  |  |
|---|---|
|  | Antracnosis (Colletotrichum sp.) |
| Maíz | ✓ Gusano barrenador (Elasmopalpus angustellus), Oruga del maíz (Heliothis armígera), Pulgón del maíz (Rhopalosiphum maidis), Roya del maíz (Puccinia sorghi), Carbón de la espiga (Sphacelotheca reiliana), Pudrición de tallo por antracnosis (Colletotrichum graminícola y Glomerella graminícola), Podredumbre de tallo y raíz (Fusarium graminearum, Gibberella zeae, Scierotium bataticola, Macrophomifla phaseoli, Diplodia maydis) |
| Pimiento | ✓ Araña Roja (Tetranychus ssp.), Podredumbre gris (Botrytis cinerea), Ceniza u Oidio, Seca o Tristeza del pimiento, Roña Sarna bacteriana, Pulgones, Trips, Mosca blanca, Heliothis |
| Pepino | ✓ Araña roja (Tretanychus urticae), Mosca minadora de las hojas del pepino, Chupado de frutos de pepino, Podredumbre blanca del cuello (Sclerotinia sclerotiorum), Virus del mosaico del pepino |
| Frijol | ✓ *Plaga de la Mosca Blanca, Chicharritas, Roya o chahuixtle (Uromyces phaseoil)*, *Moho blanco (Whetzelinia sclerotiorum), Añublo bacterial común, Trips* |

El diseño del sistema cuenta con tres capas fundamentales (Figura 1): la base de conocimiento, el motor de inferencia y la interfaz de usuario.

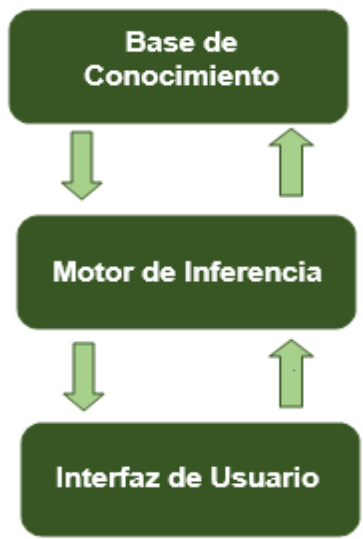

**Fig 1.** Estructura del sistema experto

**Tipografía Time New Roman 12 ptos.        Interlineado a 1.5     Justificación completa para el cuerpo**

Base de conocimientos: Contiene el conocimiento especializado que se extrae del experto en el dominio, es decir, contiene el conocimiento general sobre el dominio en el que se trabaja.

Motor de inferencia: modela el proceso de razonamiento humano, es decir, controla el proceso de razonamiento que seguirá el sistema experto, utilizando los datos que se le suministran, recorre la base de conocimiento para alcanzar una solución.

Interfaz del usuario: permite que el usuario pueda describir el problema al sistema experto, por medio de preguntas e información ofrecida.

La base de conocimiento está dividida por tipo de cultivo y para cada uno de los cultivos se cuenta con un bloque de preguntas y un bloque que muestra información sobre la plaga o enfermedad. El bloque de preguntas está formado por reglas donde se define la pregunta a realizar al usuario y la respuesta a la pregunta la cual puede ser afirmativa o negativa (Figura 2).

```
%PRINCIPAL
espregunta('es cultivo de arroz ?',X):-esrespuesta(X).
espregunta('es cultivo de tabaco  ?',X):-esrespuesta(X).
espregunta('es cultivo de tomate ?',X):-esrespuesta(X).
espregunta('es cultivo de maíz ?',X):-esrespuesta(X).
espregunta('es cultivo de pimiento ?',X):-esrespuesta(X).
espregunta('es cultivo de pepino ?',X):-esrespuesta(X).
espregunta('es cultivo de frijol ?',X):-esrespuesta(X).

principal(P1,P2,P3,P4,P5,P6,P7):-
espregunta('es cultivo de arroz ?',P1),P1='no',
espregunta('es cultivo de tabaco  ?',P2),P2='si',
espregunta('es cultivo de tomate ?',P3),P3='no',
espregunta('es cultivo de maíz ?',P4),P4='no',
espregunta('es cultivo de pimiento ?',P5),P5='no',
espregunta('es cultivo de pepino ?',P6),P6='no',
espregunta('es cultivo de frijol ?',P7),P7='no',
ph.
```

**Fig 2.** Ejemplo de reglas de un bloque de preguntas

**Tipografía Time New Roman 12 ptos.      Interlineado a 1.5     Justificación completa para el cuerpo**

Por su parte el bloque de diagnóstico contiene reglas que permiten determinar la plaga o enfermedad que afecta al cultivo, ofreciendo tanto el nombre científico como el nombre común (Figura 3).

```
tabaco(P1,P2,P3,P4,P5,P6,P7,P8,P9,P10,P11,P12):-
espregunta('se observa amarillamiento y reducción acopada de hojas en plantas jóvenes?',P1),P1='no',
espregunta('se observan grandes manchones de plantas con menor crecimiento?',P3),P3='si',
espregunta('se observa una coloración oscura en las raíces, la base del tallo, o en toda la planta?',P2),P2='no',
espregunta('se observa presencia de manchas amarillas en hojas de plantas adulta?',P4),P4='no',
espregunta('se observan hojas cloróticas, algunas con necrosis parcial en diferente grado?',P9),P9='si',
espregunta('se puede observar el sistema radicular disminuido, con raíces necrosadas, más oscuras?',P12),P12='si',
espregunta('se observa presencia de manchas de color marrón en hojas de plantas adultas?',P7),P7='no',
espregunta('se observa esporulación en hojas?',P10),P10='no',
espregunta('se observa una afectación en las raíces, que se tornan necróticas?',P5),P5='no',
espregunta('las plantas se marchitan ligeramente durante el período más caluroso del día, pero se recuperan por la noche?',P6),P6='no',
espregunta('se observa un desarrollo raquítico?',P8),P8='no',
espregunta('el sistema radicular es destruido y provoca pérdidas considerables en el cultivo?',P11),P11='no',
        pf2('C:/Programa/phytium.jpg','PYTHIUM APHANIDERMATUM  (DAMPING OFF)','C:/Programa/phytium.bmp').
```

**Fig 3.** Ejemplo de reglas de un bloque de diagnóstico de plaga o enfermedad.

Finalmente, el bloque de reglas que muestra la plaga o enfermedad que padece el cultivo, así como información de la misma y posible tratamiento (Figura 4).

```
pf2(X,Y,Z):-new(D,dialog('IMAGEN DE LA PLAGA O ENFERMEDAD')),
        mostrar2(X,D,20,40),
        new(L,label(n,'')),
        send(D, append(label(n,'PLAGA O ENFERMEDAD'))),
        send(D, append(label(n,Y))),
        send(D,append,L),
        mostrar2(Z,D,20,350),
        send(D,open).
```

**Fig 4.** Ejemplo de reglas de un bloque de información sobre la plaga o enfermedad.

**Implementación del sistema experto**

**Tipografía Time New Roman 12 ptos.         Interlineado a 1.5     Justificación completa para el cuerpo**

Para el desarrollo de este sistema experto, se utilizó Swi-Prolog Editor por ser un entorno de desarrollo para programar en prolog cuya diferencia principal con otros editores es que las consultas y la base de conocimiento están en el mismo lugar.

Se utilizó Swi-Prolog por ser un sistema basado en un lenguaje de programación que utiliza los paradigmas de programación declarativa y funcional, y que además cuenta con un motor de inferencia integrado, lo cual facilita trabajar de una manera eficiente.

SWI-Prolog es una implementación en código abierto (open source) del lenguaje de programación Prolog, licenciada bajo la GNU Lesser General Public License. Su autor principal es Jan Wielemaker, quien inició su desarrollo en 1987. Actualmente SWI-Prolog se utiliza ampliamente en la investigación y la educación, así como para aplicaciones comerciales. Una de sus ventajas es que puede ser utilizado en plataformas Unix, Windows y Macintosh. Por estas razones se seleccionó a SWI-Prolog como motor de inferencia para el desarrollo del Sistema experto.

El diseño del sistema experto cubrió los requisitos de aprendizaje rápido del manejo, evitó la entrada de datos erróneos y las preguntas y resultados se presentaron en forma comprensible para el usuario. Se desarrolló la siguiente interfaz gráfica para el sistema experto de diagnóstico de plagas y enfermedades en los cultivos de arroz, tabaco, tomate, pimiento, maíz, pepino y frijol. En las figuras 5, 6 y 7 se muestran las pantallas del sistema experto, en donde se puede ver la pantalla inicial, un cuestionario y la recomendación que el sistema arroja. En la figura 6 se presentan las preguntas realizadas para determinar el tipo de cultivo afectado. La figura 7 muestra el test de preguntas realizadas por el sistema experto al usuario para determinar la plaga o enfermedad que afecta al cultivo del arroz. En la figura 8 se muestra un cuestionario de las posibles plagas o enfermedades que afectan al cultivo del arroz y se han seleccionado en Si 4 preguntas del cuestionario para que el sistema determine cual es la plaga o enfermedad presente en la planta. La figura 9 es el resultado del diagnóstico de una de las plagas o enfermedad que pueden afectar al cultivo del arroz en la cual se muestra el tipo de plaga o enfermedad y una explicación de la misma de forma breve.

**Tipografía Time New Roman 12 ptos.    Interlineado a 1.5    Justificación completa para el cuerpo**

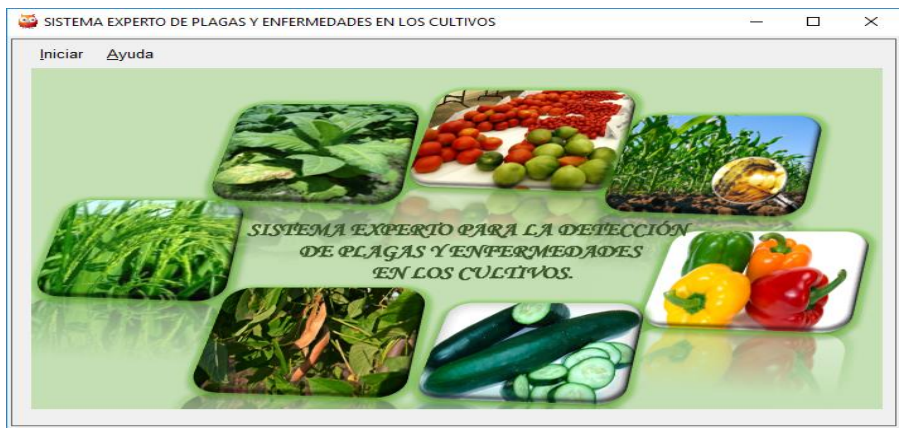

**Fig 5.** Pantalla inicial del sistema experto.

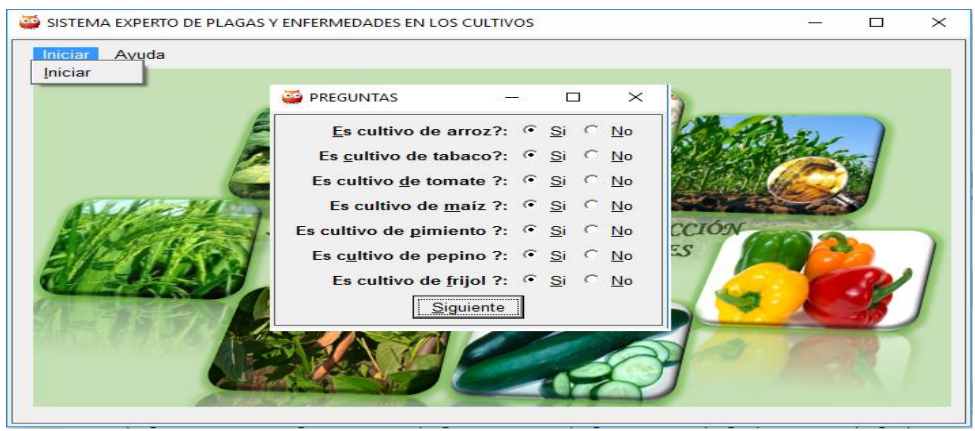

**Fig 6.** Preguntas realizadas para determinar el tipo de cultivo afectado

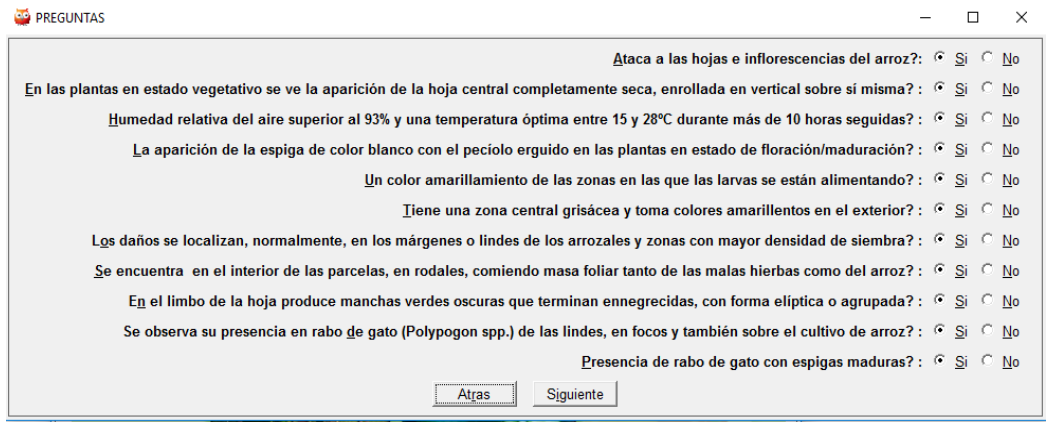

**Tipografía Time New Roman 12 ptos.        Interlineado a 1.5      Justificación completa para el cuerpo**

**Fig 7.** Test de preguntas realizadas por el sistema experto al usuario

**Fig 8.** Cuestionario que diagnostica la plaga o enfermedad pyricularia oryzae del arroz.

**Tipografía Time New Roman 12 ptos.        Interlineado a 1.5     Justificación completa para el cuerpo**

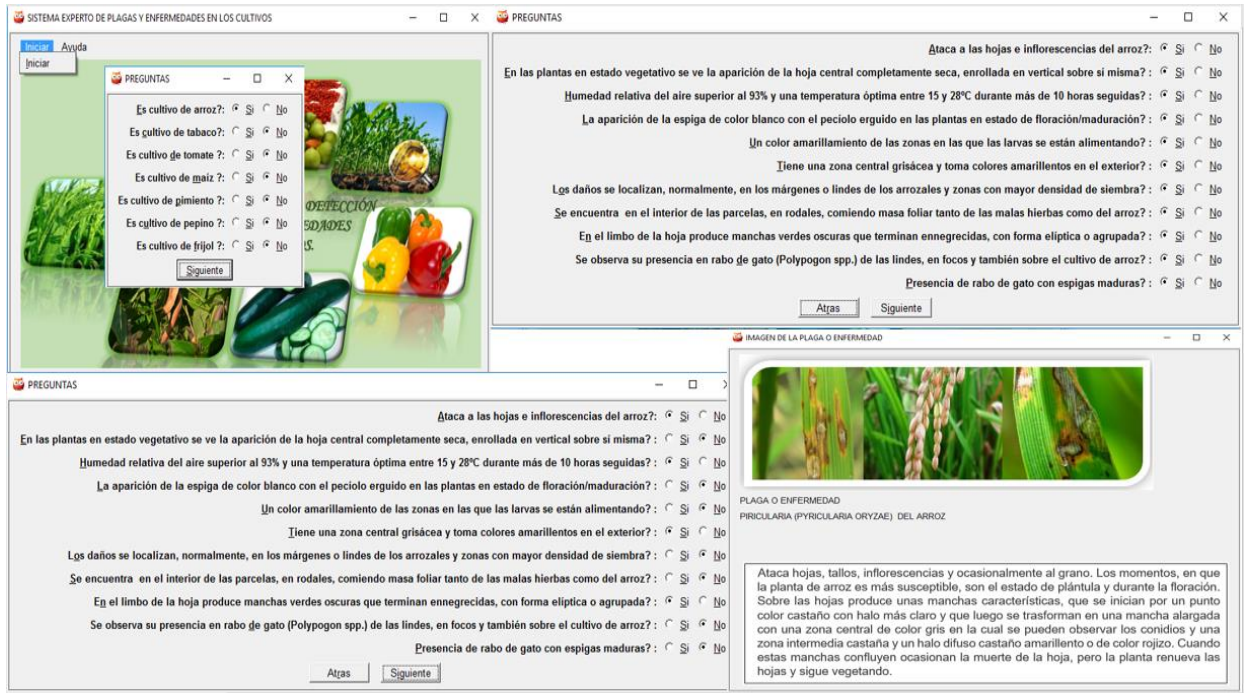

**Fig 9.** Resultado del diagnóstico de una de las plagas o enfermedad que pueden afectar al cultivo del arroz

## Conclusiones

Se logró desarrollar un sistema fácil de usar y con una interfaz amigable que interactúa con el cliente mediante preguntas, mostrando posibles curas para contrarrestar el azote de las plagas.

Se puede concluir que un sistema inteligente como el que se plantea, tiene una función de prevención y de informar a los agricultores de las posibles plagas que puedan existir en sus cultivos, dándoles así datos acerca de ellas y una posible solución a este problema.

Este sistema inteligente no solo es una herramienta de prevención, sino que brinda la información necesaria para para poder darle solución a los problemas que tiene el sector agrícola en cuanto al control de plagas.

**Tipografía Time New Roman 12 ptos.     Interlineado a 1.5     Justificación completa para el cuerpo**

Gracias a este sistema inteligente, los usuarios se verán beneficiados debido a los resultados proporcionados, mejorando así su productividad y evitando los posibles daños que afectarían sus cultivos.

Los párrafos se escribirán en Times New Roman a 12 puntos y con espaciado 1,5 y una línea en blanco como separador.

# Referencias

**Tipografía Time New Roman 12 ptos.         Interlineado a 1.5     Justificación completa para el cuerpo**

**Tipografía Time New Roman 12 ptos.     Interlineado a 1.5     Justificación completa para el cuerpo**

### Conflicto de interés

### Contribuciones de los autores

**Yosvany Medina Carbó:** contribución a la revisión bibliográfica, su análisis e interpretación. Desarrollo de la investigación. Redacción del borrador del artículo y de su versión final.

**Iracely Milagros Santana Ges:** contribución importante a la idea y diseño del estudio, revisión crítica del borrador del artículo y aprobación de la versión final a publicar.

**Saily Leo González:** contribución importante a la idea y diseño del estudio, revisión crítica del borrador del artículo y aprobación de la versión final a publicar.

### Financiación

**Tipografía Time New Roman 12 ptos.           Interlineado a 1.5      Justificación completa para el cuerpo**